# Towards Reliable and Reproducible Fetal Brain Biometry: A Deep Learning Approach Using MRI

Francesca Maccarone[a,b], Marina Di Stefano[a,c], Giorgio Longari[b], Giulia Frigerio[a], Gloria Rizzato[a], Rocco Prudentino[a], Nivedita Agarwal[a], Tommaso Ciceri[a*], Denis Peruzzo[a], Simone Melzi[b]

[a] *NeuroImaging Laboratory, Scientific Institute IRCCS Eugenio Medea, Bosisio Parini (LC), Italy,* [b] *University of Milano-Bicocca, Milan, Italy,* [c] *Image Sciences Institute, Division Imaging and Oncology, University Medical Center Utrecht, Utrecht, The Netherlands*

[*]corresponding author: tommaso.ciceri@lanostrafamiglia.it

**Abstract**

**Background and Objectives.** Fetal brain biometry is essential for quantitative assessment of brain development, supporting gestational age estimation, developmental monitoring, and detection of abnormalities. In clinical practice, measurements are manually performed, making them time-consuming and prone to variability. While automated approaches have been proposed, reproducible methods remain limited, particularly those providing anatomically interpretable landmark localization. In this study, we present a fully automated deep learning–based framework for reliable and reproducible brain biometry from 3D super-resolution–reconstructed fetal brain MRI. **Methods.** The proposed four-step pipeline derives biometric parameters by jointly estimating linear measurements and their corresponding anatomical landmarks. A 3D convolutional neural network is trained to regress landmark coordinates from brain segmentation label maps, followed by measurement-specific geometric optimization to refine landmark positions and compute measurements. The pipeline is evaluated on two publicly available fetal MRI datasets comprising 150 volumes (gestational age range: 20–37 weeks) acquired across different scanners and protocols, assessing five key biometric measurements across varying acquisition settings and providing a comprehensive evaluation of both measurement accuracy and landmark localization using quantitative metrics and visual assessment. **Results.** The framework demonstrates strong agreement with ground truth, achieving mean absolute errors below 2 mm for most measurements and mean landmark localization errors below 4 mm, with higher accuracy observed for smaller structures. Performance is consistent across datasets. Compared with the only available automated pipeline, the proposed method achieves comparable or improved accuracy for most measurements. However, corpus callosum length remains the most challenging parameter despite a targeted optimization. **Conclusions.** In conclusion, we introduce a straightforward pipeline for reliable biometry estimations, with efficiency, interpretability and scalability that support integration into clinical workflows.

**List of abbreviations**

2D: two-dimensional; 3D: three-dimensional; bBIP: Brain Biparietal Diameter; CNN: convolutional neural network; CSF: cerebrospinal fluid; DL: deep learning; GA: gestational age; GM: gray matter; GT: ground-truth; HV: height of vermis; LCC: length of corpus callosum; MRI: magnetic resonance imaging; MSL: mid-sagittal line; sBIP: skull biparietal diameter; SR: super-resolution; SVR: slice-to-volume reconstruction T2w: T2-weighted; TCD: transverse cerebellar diameter; WM: white matter

## 1. Introduction

Biometric measurements are fundamental for the quantitative assessment of fetal brain development in both clinical and research settings. In current clinical practice, these measurements are manually performed by radiologists and integrated into a comprehensive visual evaluation [1,2]. They play a key role during pregnancy by supporting gestational age (GA) estimation, monitoring fetal growth, and enabling early diagnosis and characterization of central nervous system abnormalities [2,3,4,5,6].

Ultrasound is the primary modality for prenatal screening [4,7], but magnetic resonance imaging (MRI) has gained increasing importance in fetal brain assessment over the past two decades due to its superior soft-tissue contrast [8,9,10]. Despite its advantages, MRI-based fetal brain biometry is still limited by several challenges related to both clinical practice and the imaging modality. Clinical workflows rely on a small set of linear measurements manually extracted from 2D slices aligned to specific anatomical planes, then compared with normative reference values using fetal growth centile charts. This process is time-consuming, labour-intensive, and prone to inter- and intra-observer variability [2,11,12]. Image quality is further affected by maternal and fetal motion, leading to suboptimal slice orientation, inter-slice misalignment, and inter-plane motion artifacts, particularly during the second trimester, when fetal motion is more pronounced and brain structures are smaller and more prone to measurement errors [3,9,13]. Even small inaccuracies may result in significant shifts in growth centile estimation, increasing the risk of misdiagnosis. These limitations underscore the need for more accurate, robust, and automated systems for fetal brain biometry [2,3].

Studies on automated measurements of fetal brain biometry from MRI have only recently begun to emerge [2,14], with early efforts in this direction leveraging machine learning and computer vision techniques. Initial approaches focused on direct measurement estimation from segmented images or selected slices. For example, Avisdris et al. [4] proposed a multi-step pipeline combining brain detection, reference slice selection, slice-wise and structure-wise segmentation, and measurement extraction, while She et al. [2] used whole brain segmentation followed by customized measurement-specific algorithms. Similar approaches targeted specific tasks, such as ventriculomegaly detection [15] or GA prediction using derived biometric parameters [16].

Novel methods have incorporated anatomical landmark detection to improve measurement reliability. Avisdris et al. [17] introduced a pipeline combining multiple neural networks to detect landmarks and derive measurements, while Masterl et al. [12] extended this approach to 3D super-resolution (SR) reconstructed volume using heatmap regression. Luis et al. [3] proposed a fully automated end-to-end framework based on 3D slice-to-volume reconstruction (SVR) of T2-weighted (T2w) brain volumes, atlas alignment, and 3D landmark segmentation, enabling the extraction of multiple biometric parameters and the generation of normative growth charts. However, existing methods still face important limitations, including reliance on 2D slice-based analysis, lack of standardized pipelines, limited availability of publicly accessible implementations, and lack or insufficient evaluation of landmark localization accuracy, which is critical for assessing measurement reliability. Key details of aforementioned studies are summarized in Table 1.

**Table 1**. *Overview of recent studies on automatic fetal brain biometry in MRI. The table lists the works in chronological order and reports the imaging input on which measurements are performed, the type of methodological approach (machine learning/deep learning, computer vision, geometric-prior–based, or hybrid), the type of output generated by each pipeline (direct measurement vs landmark-based measurement) and, in the last column, the extracted biometric parameters.*

| Study [reference] | Input data type | Method approach | Output type | Extracted metrics |
|---|---|---|---|---|
| [17] | 2D slices | DL | landmark-based linear measurements | brain biparietal diameter (bBIP), skull biparietal diameter (sBIP) and trans-cerebellum diameter (TCD) |
| [4] | 2D slices | DL + geometric prior | linear measurements | cerebral biparietal diameter (CBD), bone biparietal diameter (BBD) and TCD |
| [2] | 2D slices | DL + geometric prior | linear measurements | CBD (referred as CBPD in the original text), TCD and left and right atrial diameters (LAD and RAD) |
| [15] | 3D reconstructions | DL + geometric prior | linear measurements | size of lateral ventricles |
| [16] | 2D slices | DL + CV | linear measurements for GA estimation | biparietal diameter (BPD) and fronto-occipital diameter (FOD) (i.e. width and length of brain, respectively), and head circumference (HC) |
| [12] | 3D SR reconstructions | DL | landmark-based linear measurements | 11 standard biometric measures |
| [3] | 3D SR reconstructions | DL | landmark-based linear measurements | 13 standard biometric measurements, regarding regarding skull, brain ventricles, corpus corpus callosum, cerebellum, pons and vermis |

*DL: deep learning, CV = computer vision, SR = super-resolution, GA = gestational age.*

In this work, we address these limitations by proposing a deep learning (DL)-based framework for automated fetal brain biometry from 3D MRI. The method combines landmark regression with a geometric refinement step to accurately localize anatomical landmarks and derive corresponding measurements from SR T2w fetal brain volumes. We evaluate our pipeline on five key biometric measurements across different acquisition settings, using two publicly available fetal datasets, and provide a comprehensive evaluation of both measurement accuracy and landmark localization, with quantitative metrics and visual assessment. In addition, we compare our method with the only publicly available landmark-based pipeline for fetal MRI brain biometry and release our implementation to the scientific community.

## 2. Methods

### 2.1 Data

#### 2.1.1 Datasets

We used two publicly available datasets: the Zurich dataset from the 2024 FeTA Challenge [18,19] and the Developing Human Connectome Project (dHCP) dataset [20]. Both include fetal brain 3D reconstructed images from T2w MRI acquisitions.

*"Zurich dataset"*: it comprises fetal brain data collected at Universitäts Kinderspital Zürich. For this study, we start from 80 SR T2w images: 40 volumes were reconstructed using MialSRTK [21,22], with an isotropic resolution ranging from 0.43 to 0.7 mm, while the remaining images were reconstructed via the Image Registration Toolkit (IRTK) [23], with a final isotropic resolution of 0.5 mm. All volumes were subsequently zero-padded to a uniform size of 256x256x256 voxels. Due to image noise limiting manual annotation (Fig. S1 in Supplementary), subjects without ground-truth landmarks are excluded, while those with partial annotations are retained, resulting in a final cohort of 70 subjects (31 neurotypical, 39 pathological). This dataset is available on Synapse [24].

"*dHCP dataset*": we consider a subgroup of 80 subjects from the dHCP-4$^{th}$ release fetal dataset, sampling uniformly across GAs (Fig. 1). Final data are isotropic 3D volumes reconstructed using a SVR method, with automatic rejection of corrupted data [25].

The distribution of samples from both groups is represented in Fig. 1, while Table 2 reports the main information regarding the datasets.

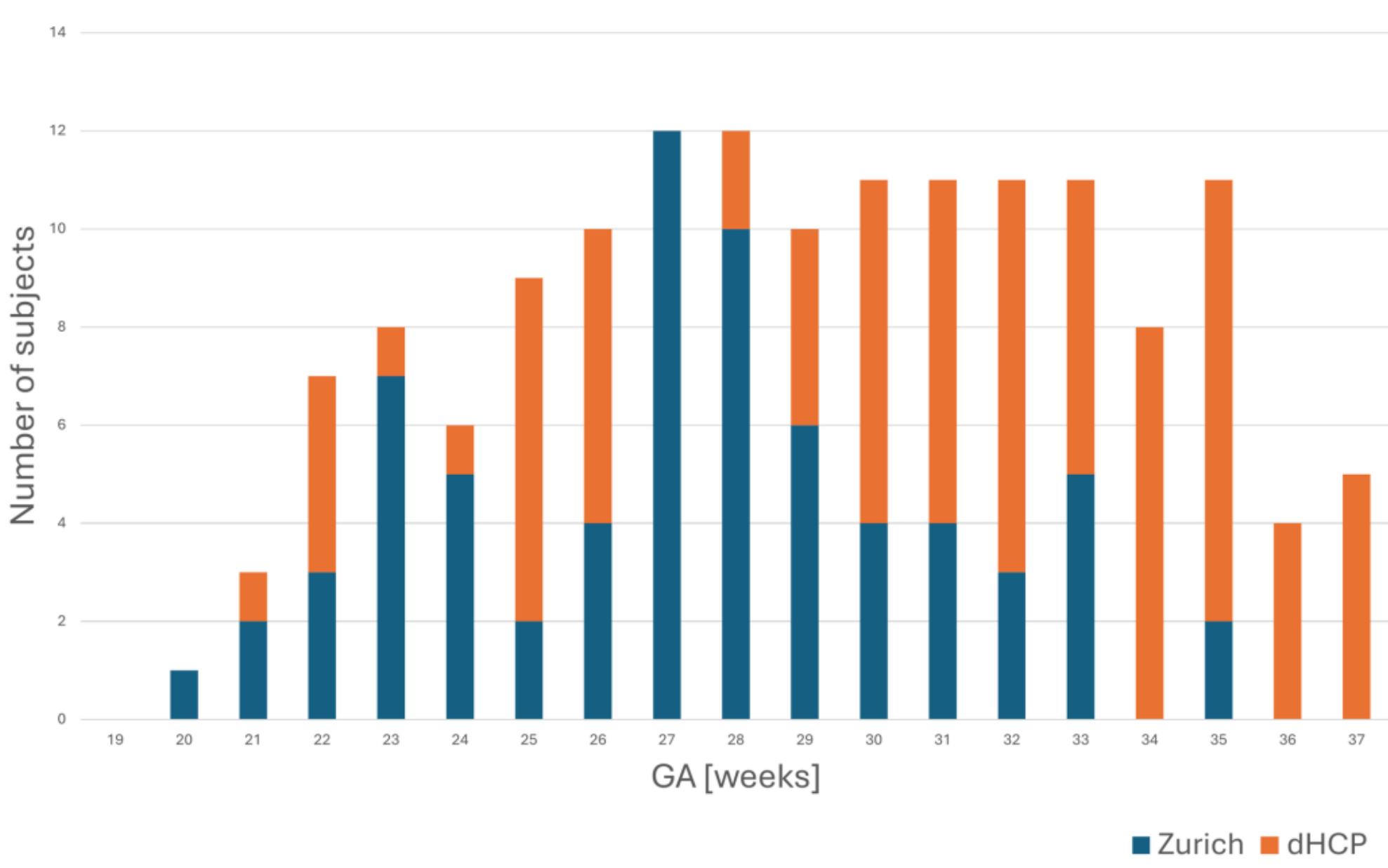

**Fig. 1.** *Distribution of subjects across gestational ages for the two datasets: Zurich and dHCP. For each GA, bars represent the cumulative number of subjects in both datasets splitted into subjects from the Zurich dataset (blue) and from the dHCP*

*dataset (orange).*
*GA: gestational age.*

**Table 2.** *Main acquisition parameters and population statistics for the datasets used in the study, namely the Zurich and dHCP datasets.*

| | ***Zurich dataset*** | ***dHCP dataset*** |
|---|---|---|
| *Encoding* | FSE | TSE |
| *magnetic strength [T]* | 1.5, 3 | 3 |
| *manufacturer (model)* | GE (Signa Discovery MR450 and MR750) | Philips (Achieva) |
| *n° of coil channels* | 8, whole-body coil | 32, cardiac coil |
| *acquisition resolution [mm$^2$]* | 0.5 x 0.5 | 1.1 x 1.1 |
| *slice thickness [mm]* | 3 ÷ 5 | 2.2 |
| *TE [msec]* | 120 (minimum) | 250 |
| *TR [msec]* | 2000 ÷ 3500 | 2265 |
| *n° subjects* | 70 | 80 |
| *GA (mean ± SD) [weeks]* | 27.03 ± 3.6 | 30.07 ± 4.4 |
| *GA range [weeks]* | 20 – 34.8 | 21 – 37 |

*FSE: fast spin echo; TSE: turbo spin echo; TE: echo time; TR: repetition time; GA: gestational age.*

### 2.1.2 Tissue labels

The proposed framework is based on label maps, i.e. segmentation outputs in which each voxel is assigned a tissue label. We use label maps from different sources (i.e. manual and automatically derived ones) to investigate the impact of the segmentation on the biometric measurements.

In the Zurich dataset, manually defined label maps include 8 labels: background and non-brain tissue (BG), extra-axial cerebrospinal fluid (CSF), gray matter and developing cortical plate (GM), white matter and subplate (WM), lateral ventricles (LV), cerebellum (CBM), deep gray matter (DGM, including thalamus and putamen) and brainstem (BS). In the dHCP dataset, label maps were automatically derived using DrawEM [26,27] and include 9 labels, introducing the hippocampi tissue to the ones already listed for the Zurich dataset. For consistency, we merged hippocampi with DGM mask to obtain 8-tissues label maps.

We also derived label maps using BOUNTI [28,29], which produces a label map with 19 different tissues. We combined them (Table S1 in Supplementary) to derive a label map matching the previously described one.

### 2.1.3 Landmark manual placement

In this setting, landmarks are fiducial anatomical points from which the biometric measures are derived.
The measurements under investigation are: 1) Length of Corpus Callosum (LCC), 2) Height of Vermis (HV), 3) Transverse Cerebellar Diameter (TCD), 4) Brain Biparietal Diameter (bBIP), and 5) Skull Biparietal Diameter (sBIP). Each measurement is defined on a specific anatomical plane according to established clinical guidelines [30] and represented in Fig. 2.

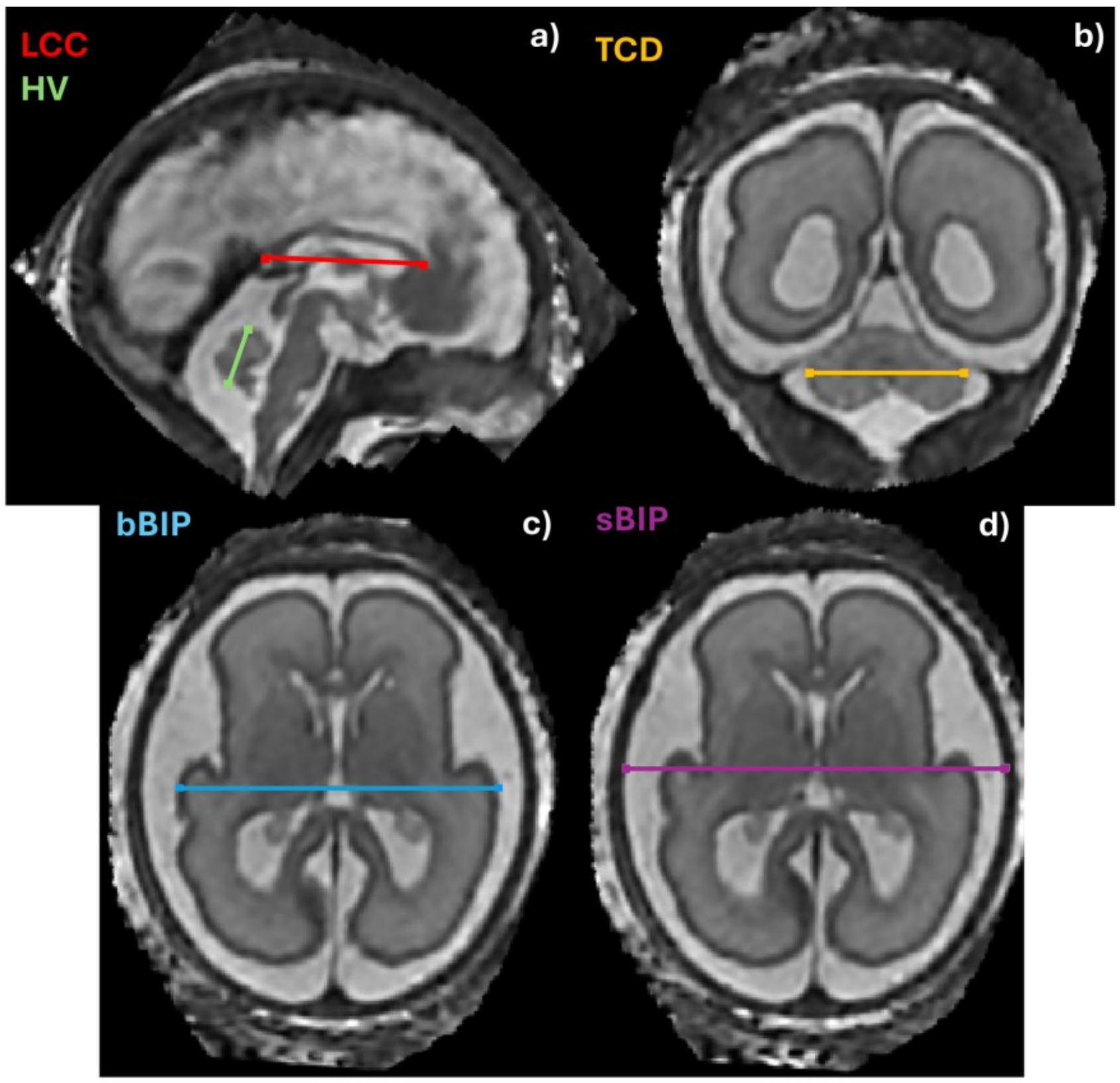

**Fig. 2.** *Example of the selected five biometrical measures. For each measure, corresponding landmarks manually placed on a 27-week GA subject from the Zurich dataset are shown. Panel (a) reports HV (green) and LCC (red) in sagittal view; panel (b) reports TCD (orange) in coronal view; panel (c) and (d) report bBIP (cyan) and sBIP (magenta), respectively, in the axial plane.*
*bBIP: biparietal brain diameter; HV: height of the vermis; LCC: length of corpus callosum; sBIP: skull biparietal diameter; TCD: transcerebellar diameter.*

The Zurich dataset includes expert-defined landmarks following the 'Fetal biometry annotation guidelines' [31], adapted from [30] to standardize 2D biometry measurements. The dHCP dataset does not include landmark annotation, thus we manually labelled data following [31]. In detail, two trained raters independently labelled the images under expert neuroradiologist supervision. At the end of the annotation process, each volume includes five pairs of landmarks from which ground truth biometric measurements are computed, consistently with the Zurich dataset.

### 2.2 Pipeline workflow and experiments

Our pipeline consists of four main steps, which are described in the next subsections and illustrated in Fig. 3.

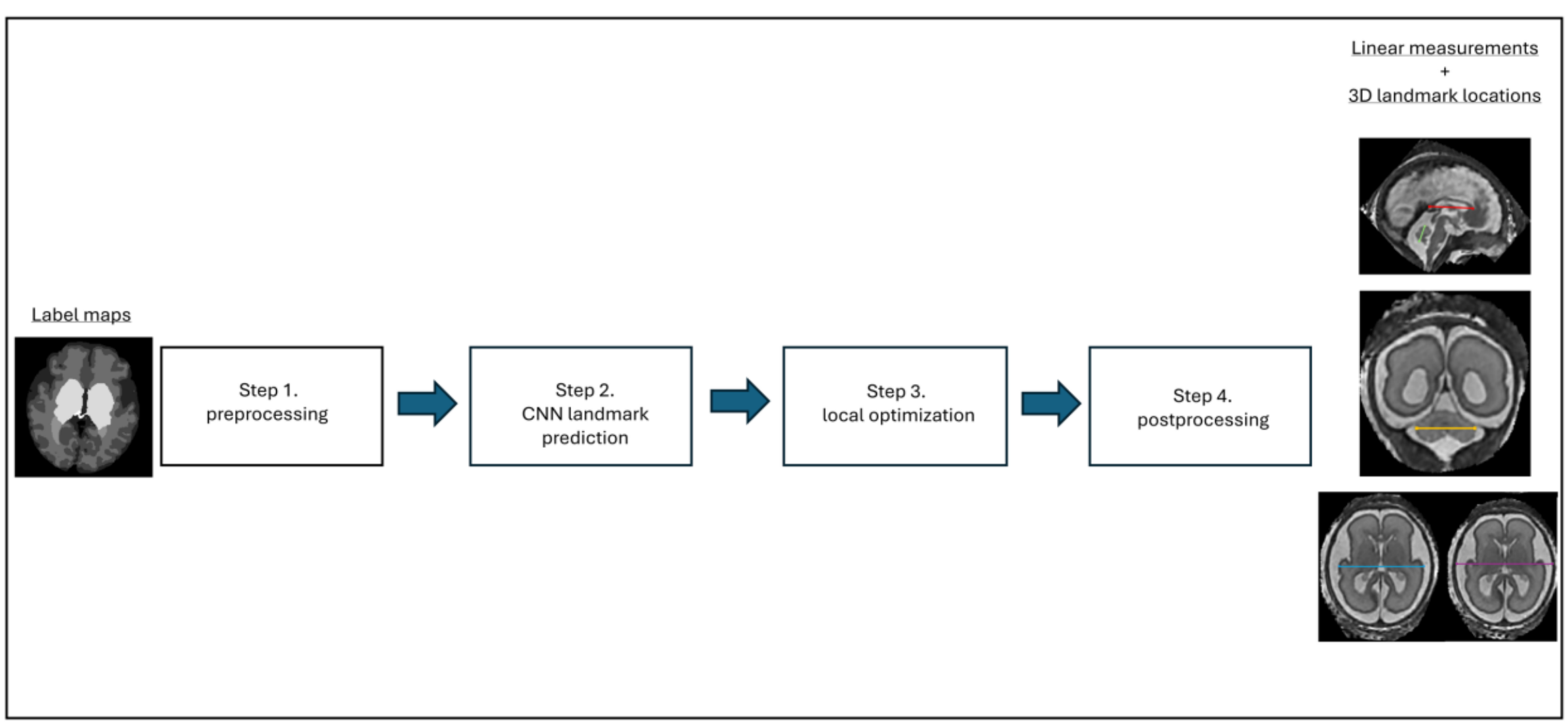

**Fig. 3.** *Overview of the proposed four-step fully automatic pipeline. The pipeline takes as input label maps derived from T2w MRI and produces, as final outputs, linear biometric measurements together with the corresponding 3D landmark positions from which the measurements are computed. In the first step, preprocessing makes raw acquisitions, their corresponding label maps and the anatomical landmarks suitable for model training (subsection 2.2.1). Building upon [42], in the second step a CNN predicts coarse landmark positions (subsection 2.2.2), while the third step refines these positions (subsection 2.2.3). The final step computes the biometric measurements from the refined landmarks (subsection 2.2.4).*

### 2.2.1 Step 1: Preprocessing

All raw T2w images and label maps (from the datasets or derived with BOUNTI) are rigidly registered to the Uus atlas [32] at the 36th gestational week using the Advanced Normalization Tools [33,34], ensuring consistent alignment in a standard space regardless of the SR method or voxel resolution. Volumes containing label maps are resized to a target shape of 128x128x128 using nearest-neighbour interpolation (0.5 mm isotropic), to match the model input dimensions.

For landmarks, Zurich original volumes are aligned to a 'measure space', where the SR images were manually aligned to a standard orientation for annotation. We map these landmarks back to the subject's native space using the provided transformations, whereas dHCP landmarks are already defined in native space. At this stage, all volumes containing landmarks are registered to the Uus atlas, consistent with T2w preprocessing. To avoid losing landmarks during resampling, we dilate each of them by applying a morphological operation using a structuring element, whose size is defined by the k parameter, prior to registration.

### 2.2.2 Step 2: CNN for landmark prediction

In the original formulation [35], a 2D CNN is used to detect isocentres for radiotherapy applications. We adapt this approach by training a 3D CNN to regress the coordinates of the landmarks starting from the label maps. The CNN takes 128x128x128 input images and consists of four convolutional layers for feature extraction, two max pooling layers applied in the last two layers to downsample the feature maps and summarize the information, and a fully connected head to perform the landmarks regression. A default kernel of size 3x3x3 is utilized.

*Training settings*

To assess the impact of label quality on model learning, we train our model considering two main settings. In the first configuration, the model is trained using the ground-truth label maps provided with the datasets (hereafter '*GT*'). In the second configuration, the same architecture is trained using automatically generated label maps from BOUNTI (hereafter '*BOUNTI*'). This experimental design enables direct evaluation of how the quality and potential noise in the training labels influence the model's ability to learn geometric representations and predict reliable biometric measurements.

*Training strategy*

In our experiments, no data augmentation is applied. The two datasets are combined into a single collection, which is split into train, validation and test sets following the conventional 80%-10%-10% proportion. Data are divided into the three subsets while preserving the relative proportions between the two datasets across different GAs (Fig. S2 in Supplementary), with a fixed random seed to ensure reproducibility.

Training uses Adam optimizer with randomly initialized weights and a learning rate of 5e-5, which is reduced by a factor of 10 if the MSE training loss does not decrease for 5 consecutive epochs. No cross-validation is applied but we consider 10% of the samples for validation during hyper-parameter tuning. The model is trained with a batch size of 16 for up to 300 epochs, with an early stopping criterion that monitors the loss calculated on the validation set with a patience of 10 epochs. The model is trained until performance plateau, and the best weights are retained.

### 2.2.3 Step 3: Landmark local optimization

The third step implements a local optimization strategy to refine landmark pair localization. For measurement whose landmarks lie within a specific anatomical structure in the label map (e.g. cerebellum for HV and TCD), the optimization maximizes the distance between the two points along the line defined by the initial CNN predictions, while constraining the optimized coordinates to remain within the corresponding structure. The procedure is measurement-specific and requires the associated anatomical mask. This approach is applied to 4 of the 5 measurements, while LCC is handled differently due to the absence of the corpus callosum segmentation label.

The general optimization proceeds as follow:

1. A line is defined between the two coarse landmarks predicted by the CNN along the anatomical axis relevant for the measurement. Landmarks are projected onto the corresponding anatomical plane by considering only the coordinates defining that plane (i.e., axial for bBIP and sBIP, mid-sagittal for LCC and HV, and coronal for TCD) and the line is then estimated between these points via least-squares fitting.

2. Discrete points are sampled along the computed line in image space and filtered to retain only those within the measurement-specific structure in the label map (e.g. for bBIP, whose actual landmarks lay on the border of GM, we retain only points within the GM label).

3. The pair of points with maximum Euclidean distance is selected as the final landmark estimate.

*Corpus callosum optimization*

Since for LCC landmarks lie along the corpus callosum border, which does not have a distinct label in our label maps, we adopt a dedicated strategy using the WM mask as a proxy for the corpus callosum structure and enforcing anatomical constraints by restricting landmarks to interfaces between WM and GM and between GM and CSF.

Starting from the CNN-predicted landmarks, the mid-sagittal plane is identified as the midpoint between them along the orthogonal axis. To account for minor variability in landmark localization, a neighbourhood of five slices (offsets from −2 to +2) is considered. In each slice, the initial landmark coordinates are projected onto WM-GM and WM-CSF interfaces, obtained by dilating the corresponding masks and computing their intersection. Each landmark is snapped to the nearest interface voxel by minimizing the Euclidean distance. Among all candidates points across slices, the pair maximizing the Euclidean distance is selected as the refined landmark pair estimate. This approach (Fig. 4) enforces anatomical plausibility while capturing the full extent of the structure of interest despite the absence of a dedicated corpus callosum label.

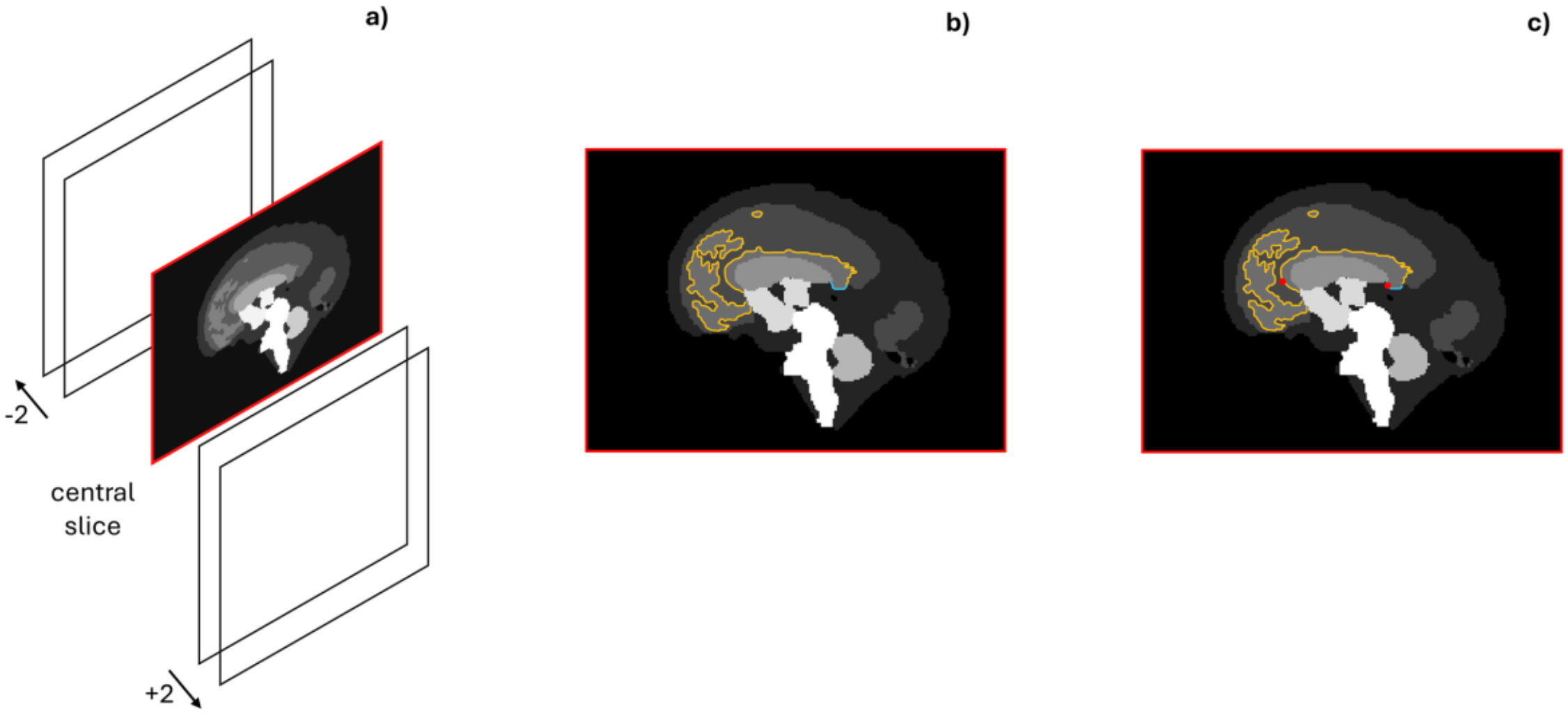

**Fig. 4.** *Salient steps of the optimization procedure for automatic corpus callosum length extraction. This strategy includes a) central sagittal slice identification and neighbourhood localization (+2 and -2 offset); b) white matter-gray matter (yellow lines) and white matter-cerebrospinal fluid (blue lines) interfaces identification; c) landmarks (red dots) projection to the nearest interface.*

### 2.2.4 Step 4: Postprocessing

Finally, the estimated landmarks, which are defined in the atlas space [32], are moved back to the subject's space. Estimated landmarks are first encoded as a sparse 3D label volume at atlas resolution, where each landmark is assigned a unique label (i.e. the corresponding measurement). Dilation is applied to preserve landmarks during transformation. The volume is then transformed to the original space using the transformation from the preprocessing step. Voxels corresponding to each label are clustered to account for interpolation-induced dispersion and each landmark position is taken as the centroid of its cluster, yielding a single 3D coordinate in the native space. The biometric measurements are then measured as Euclidean distance between corresponding landmark pairs.

### 2.3 Evaluation and Metrics computation

Inter-observer reproducibility is evaluated on a subset of 10 randomly selected subjects from the dHCP dataset using the Intraclass Correlation Coefficient (ICC), specifically ICC3, to assess the agreement in landmark placement between the two raters [36].

We perform a quantitative error analysis of both the automatically derived measurements and the predicted landmark positions on test set, using manual annotations as ground truth. We evaluate two common metrics used in biometry: the absolute error and percentage error.

To assess the accuracy of the predicted measurements, we compute some statistical measures, namely the mean absolute error (MAE), root mean squared error (RMSE) and the Pearson correlation coefficient (R), where close-to-1 values indicate strong agreement between predictions and reference values.

To evaluate the performance of our pipeline, we compare it with the method by [3], using their docker-based script (svrtk) [37] on our test set to obtain the corresponding biometric measurements, with the analysis limited to the measurements available from both pipelines. To determine whether the two methods differ significantly, we apply a paired Wilcoxon signed-rank test to the distributions of absolute errors for each measurement and calculate the effect size using Cliff's delta [38] to quantify the magnitude of any performance differences between the methods.

## 3. Results

As the two segmentation methods do not yield statistically significant differences (see Supplementary section S1), the following results report only those obtained using GT label maps as input to the pipeline for conciseness. Experiments with BOUNTI segmentations are nonetheless performed and the results are reported in Supplementary S1 section.

We also evaluate the effect of lowering the dilation parameter k (see subsection 2.2.1), reducing it from an initial value heuristically set to 5 to 3, which represents the minimum value ensuring detection of all landmarks. This value is selected as the optimal parameter for all subsequent analyses (see Supplementary section S2).

### 3.1 Inter-observer agreement

All ICC estimates exceed 0.922, indicating an excellent inter-rater agreement [36] (Table 3).

**Table 3.** *Inter-rater agreement results for dHCP dataset. The intra-class coefficient (ICC3), together with the corresponding p-value and Confidence Interval at 95th percentile (CI95%), are reported for each linear biometric measurement.*

| ***Measurement*** | ***ICC3*** | ***p-value*** | ***CI95%*** |
|---|---|---|---|
| *LCC* | 0.922 | 2.60E-05 | [0.72, 0.98] |
| *HV* | 0.959 | 2.00E-06 | [0.85, 0.99] |
| *bBIP* | 0.979 | 6.99E-08 | [0.92, 0.99] |
| *sBIP* | 0.997 | 5.02E-12 | [0.99, 1.0] |
| *TCD* | 0.992 | 7.09E-10 | [0.97, 1.0] |

### 3.2 Biometric measurements analysis

Fig. 5 shows absolute and percentage error distributions for each measurement across the entire test set (upper row) and the Zurich and dHCP subsets (bottom row). Paired t-tests comparing the predicted and GT values on the full test set show a significant difference only for LCC ($p = 0.0003$), indicating underestimation by the proposed pipeline, whereas all other measurements are not significantly different ($p>0.05$). After stratifying the test set by dataset, we perform paired t-tests to compare automatic measurements with the corresponding GT values within each subgroup. Consistent with the overall analysis, a statistically significant difference is observed only for LCC (Zurich: $p = 5.98\times10^{-5}$; dHCP: $p = 0.002$). Table 5 summarizes error metrics (MAE and RMSE) and R for our results ('ours'). Consistent with the error analysis in Fig. 4, bBIP, sBIP, and TCD exhibit comparable accuracy, while LCC exhibits higher error. Correlation values are consistently high ($R = 0.85$–$0.99$), indicating a strong to very strong linear relationship between the reference and the predicted values, supporting the reliability of the method.

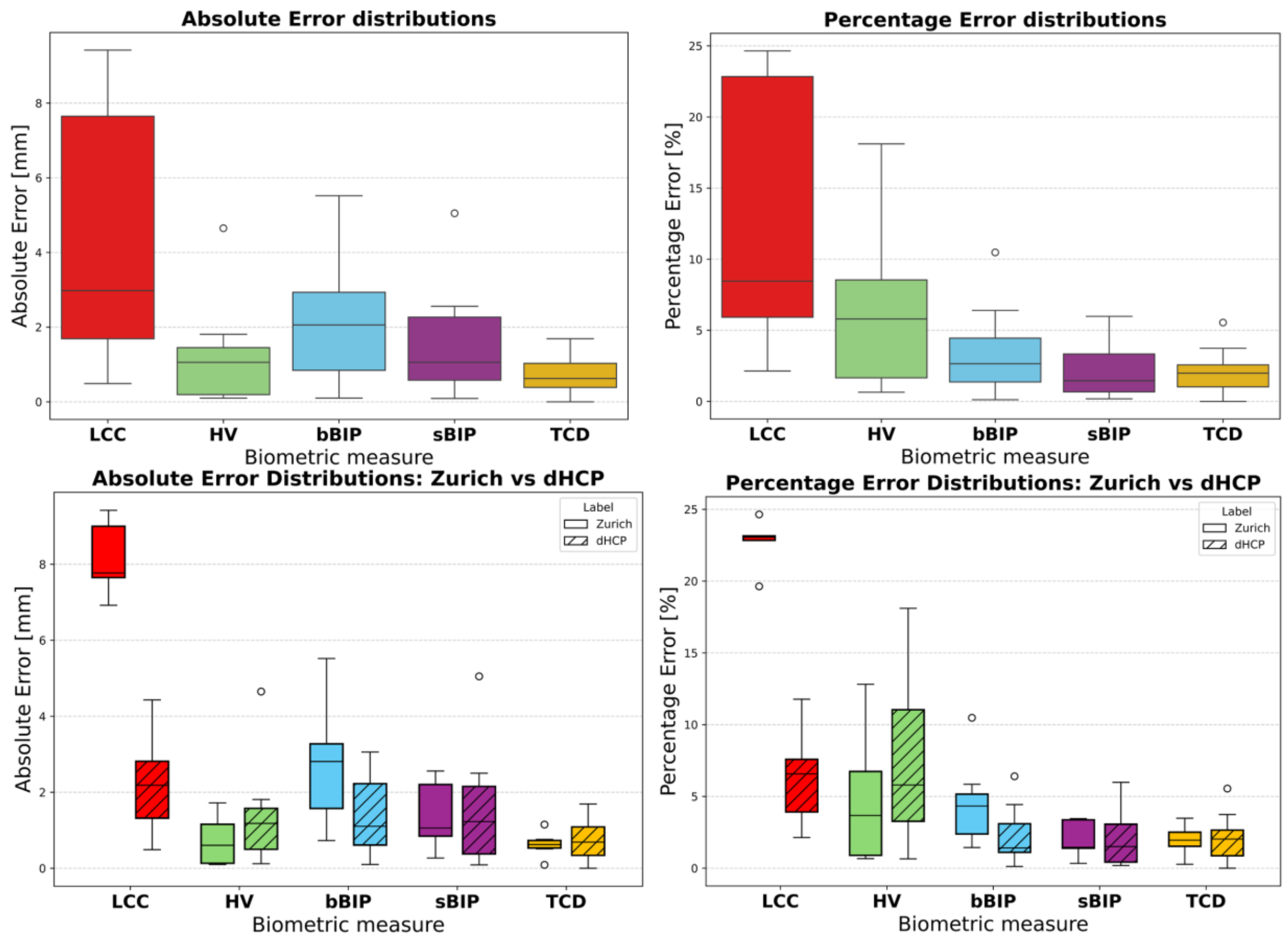

**Fig. 5.** *Error distributions across test set subjects. The top row reports the absolute and percentage error distributions for each biometric measurement. The bottom row reports the absolute and percentage error distributions subdivided according to the datasets: for each measure, the left box refers to the Zurich dataset, while the right to the dHCP dataset.*

**Table 5.** *Landmark localization distance error. Distance between the ground truth and the predicted landmarks is reported as the mean ± standard deviation. 'Landmark1' and 'Landmarks2' indicate the pair of landmarks used to derive each measurement. All values are expressed in millimetres (mm).*

| ***Measurement*** | ***Landmark1 distance error [mm]*** | ***Landmark2 distance error [mm]*** |
|---|---|---|
| *LCC* | *4.04 ± 3.08* | *4.26 ± 2.56* |
| *HV* | *2.68 ± 1.44* | *2.44 ± 1.67* |
| *bBIP* | *6.00 ± 3.69* | *4.89 ± 3.07* |
| *sBIP* | *5.42 ± 3.07* | *4.62 ± 1.89* |
| *TCD* | *2.53 ± 1.47* | *2.43 ± 0.93* |

Table 4 also reports error metrics for the competitor method by [3] ('Luis") across the test set. Our approach ('ours') generally yields slightly or consistently lower error, except for LCC. Paired Wilcoxon signed-rank tests show no significant differences for HV (p = 0.5935), bBIP (p = 0.3028), and TCD (p = 0.263) and the Cliff's delta coefficients reveal only small improvements in measurement accuracy for our method (Cliff's delta: 0.143, 0.279, 0.243, respectively). Significant differences are observed for LCC (p = 0.0007), where the competitor method demonstrates a notable improvement in accuracy (Cliff's delta = 0.843), and for sBIP (p = 0.0001), where our method achieves a large improvement (Cliff's delta = 0.821).

**Table 4.** *For each measurement, the second and third rows report the mean absolute error (MAE) and root mean squared error (RMSE) computed across the test set for both our pipeline ('ours') and the competing method proposed by Luis et al. [3]. Additionally, the third row includes the Pearson correlation coefficient (R). The best results are shown in bold.*

| | *LCC* | *HV* | *bBIP* | *sBIP* | *TCD* |
|---|---|---|---|---|---|
| | *ours / Luis'* | *ours/ Luis'* | *ours /Luis'* | *ours / Luis'* | *ours /Luis'* |
| *MAE [mm]* | 4.47 / **1.77** | **1.12** / 1.39 | **2** / 2.17 | **1.52** / 4.4 | **0.69** / 0.77 |
| *RMSE [mm]* | 5.45 / **2.15** | **1.6** / 2.13 | **2.4** / 2.49 | **1.98** / 4.55 | **0.82** / 0.89 |
| *Pearson R* | 0.85 / **0.92** | **0.94** / 0.92 | 0.98 / 0.98 | 0.99 / 0.99 | 0.96 / **0.99** |

The larger error distribution associated with LCC suggests that the specific optimization procedure may be suboptimal. To investigate this, we compare LCC estimates with and without the optimization step. The optimization process helps improving the accuracy, lowering the median value of both absolute and percentage error distributions (Fig. 6). Paired t-tests comparing errors with and without the optimization shows a significant difference in favour of the estimates after optimization (absolute error: p-value = 0.003; percentage error: p-value = 0.002).

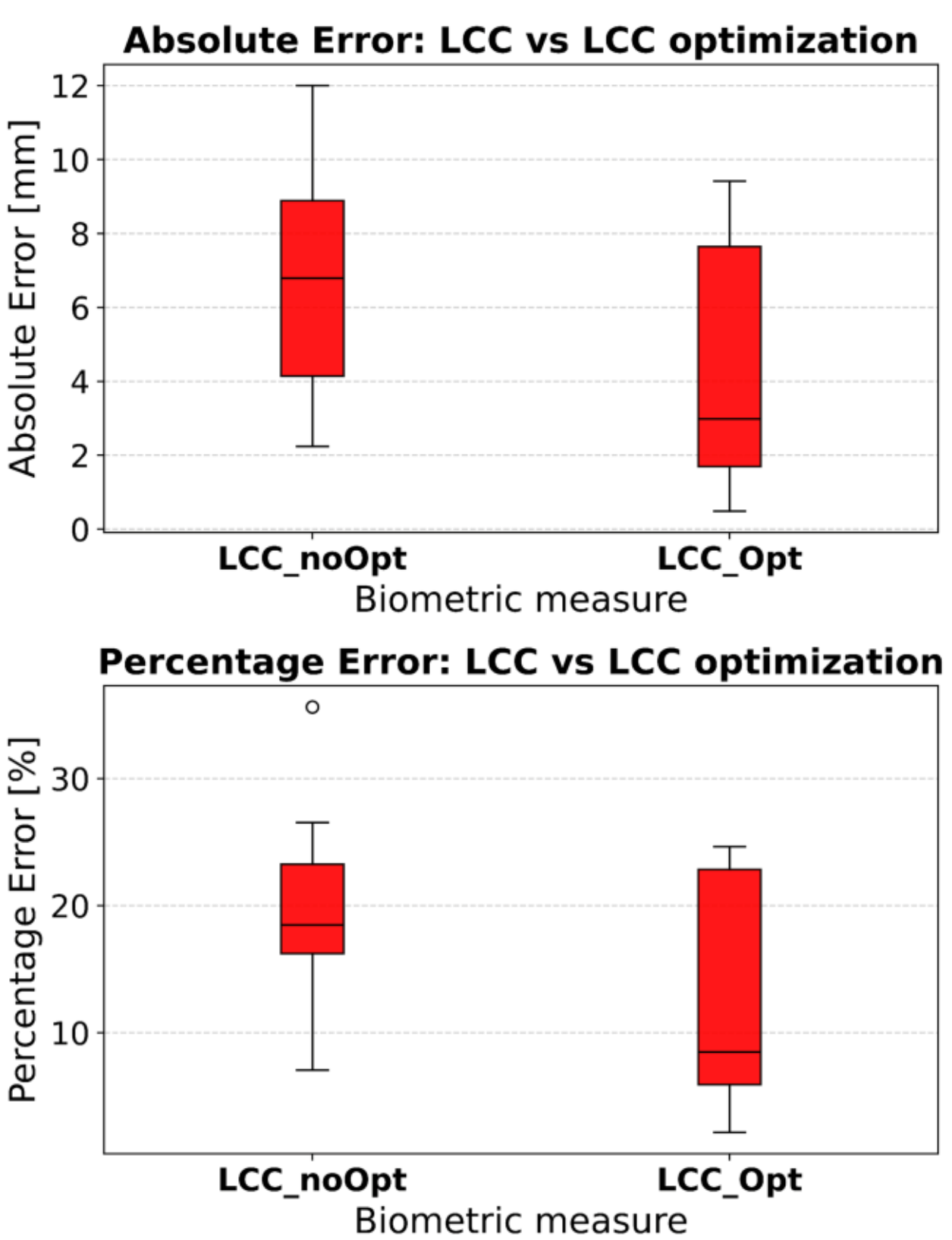

**Fig. 6.** *LCC results***.** *Absolute error (in millimetres), and percentage error distributions in the test set for the LCC measurement with (LCC_opt) and without (LCC_noOpt) the application of the ad-hoc optimization process to the measurement landmarks (see 'Corpus callosum optimization' in subsection 2.2.3).*
*Acronym: Opt= optimization*

### 3.3. Landmarks analysis

We complement the measurement error analysis by evaluating the accuracy of the estimated landmark locations used to derive final measurements (Table 5). The absolute error (i.e. the Euclidean distance between predicted and GT landmarks) is reported in Fig. 7, both in 3D space (left panel) and decomposed along individual coordinate axes (right panel) to identify landmark-specific error patterns.

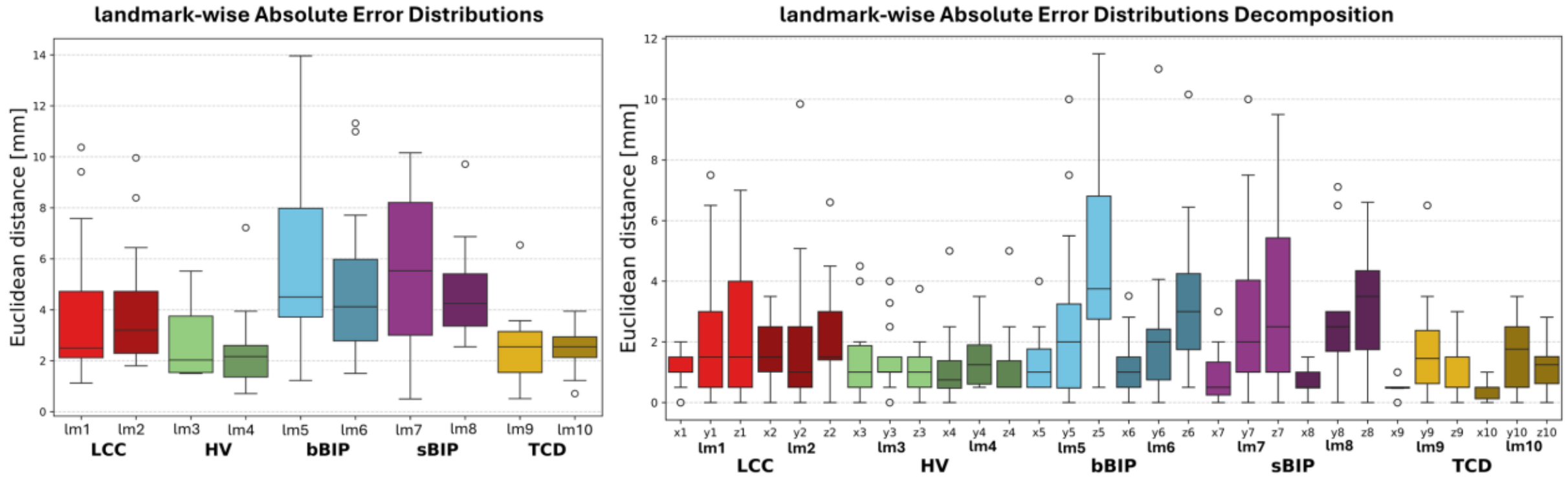

**Fig. 7.** *Distribution of the absolute error in the landmark location estimate. The left panel reports for each measurement the distance between the estimated landmark and the ground truth computed in the 3D space. The right panel decomposes the distance along the three axes (i.e. xyz) following the standard MRI convention (i.e. x along the left-right direction, y along the rear-front direction, z along the feet-head direction). In both panels, the endpoints referring to the same measurement*

*are depicted with different shades of the same colour.*
*Acronym: lm =landmark*

Fig. 8 shows representative cases for each measurement, illustrating predicted and GT landmark locations for subjects with median and maximum distance error. This representation enables direct inspection of deviations from manual annotations.

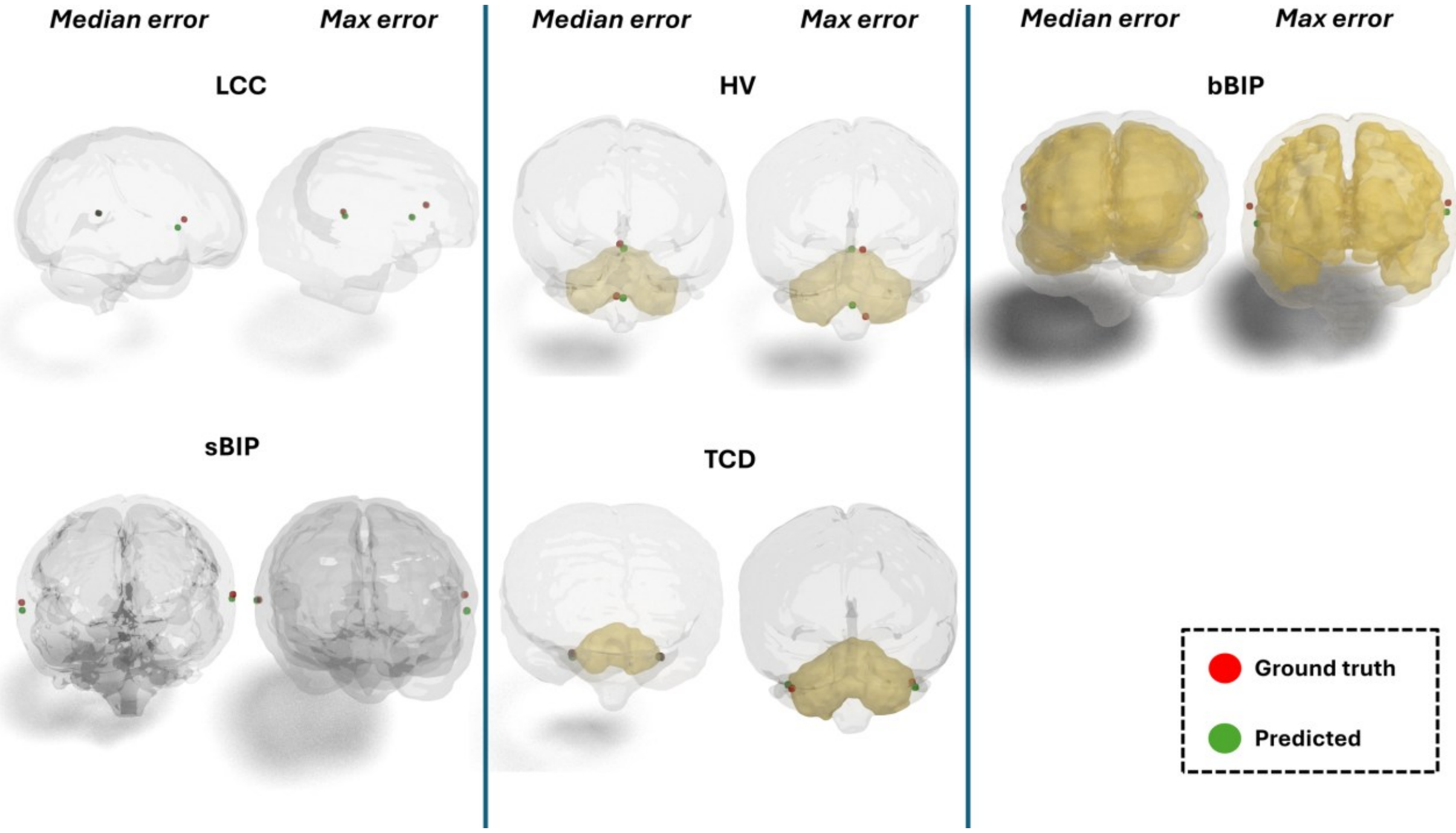

**Fig. 8.** *Visual representation of estimated and ground-truth landmark locations. For each measurement, the subjects exhibiting the median and maximum error are reported on the left and on the right, respectively. Red dots refer to the ground-truth landmark positions while green dots refer to the positions predicted by our method. The landmarks representation has been dilated and all depicted surfaces reduced to unitary scale just for visualization purposes.*

## 4. Discussion

This study proposes a DL-based pipeline for fully automated brain biometry in fetal MRI, designed to operate robustly across imaging sites, acquisition protocols, and GAs. Starting from label maps derived from T2w volumes, a CNN regresses the 3D coordinates of landmark pairs, which are subsequently refined through an optimization procedure. A dedicated strategy is introduced for the corpus callosum, which is not explicitly included in the segmentation maps. The pipeline outputs both biometric measurements and corresponding 3D landmark locations, enabling intuitive validation and increasing confidence in automated analysis. We consider five clinically relevant biometric parameters routinely used in fetal brain assessment. Among these, sBIP, bBIP, and TCD are widely adopted in ultrasound and MRI-based studies [2,17,39]. In addition, LCC and HV are included due to their relevance in monitoring fetal development and detecting brain malformations, including agenesis of the corpus callosum and cerebellar or vermian anomalies [40,41].

To improve robustness and generalizability, we consider two publicly available datasets with different characteristics. Measurement error analysis (Fig. 5) highlights differences in precision across parameters. In particular, bBIP, sBIP, and TCD show compact absolute and relative error distributions that remain stable across datasets, suggesting that their accuracy is primarily limited by intrinsic methodological factors rather than dataset-specific variability. In contrast, HV exhibits low absolute errors but higher relative errors, reflecting its small anatomical size, where small millimetre deviations lead to large proportional errors. This behavior is consistent with known limitations of fetal MRI resolution and partial-volume effects, which disproportionately affect small structures [40]. Overall, absolute errors are consistent with previously reported automated methods [2,4,11,12] and fall within the variability observed at manual level, except for LCC. Comparison with the method by [3] shows that our approach achieves comparable performance across several parameters, including HV, bBIP, and TCD, and outperforms it for sBIP (Table 5). Both methods demonstrate good accuracy for small structures such as the cerebellar vermis, highlighting the effectiveness of 3D SR reconstructions in preserving fine anatomical details. However, the competitor achieves substantially more accurate LCC measurements.

LCC represents the most challenging parameter in this study, showing the lowest ICC in manual annotation (Table 4), the largest absolute and relative errors, and the only clear dataset-dependent differences. This variability likely reflects multiple factors, including optimization complexity, low anatomical contrast, and sensitivity to inconsistencies in manual landmark placement. These findings are consistent with previous reports highlighting the difficulty of identifying the corpus callosum in fetal T2w MRI, where this thin midline structure may be poorly visualized due to acquisition factors and incomplete development during early gestation [5]. Reconstruction-related effects, such as partial-volume artifacts and boundary blurring, further exacerbate these challenges. Prior studies [42,43] also emphasize that the small size of midline structures reduces measurement reliability and that correct mid-sagittal plane selection is critical for consistent identification of the genu and splenium. Although suboptimal compared with the general optimization strategy, our ad-hoc optimization procedure for LCC yields measurable improvements, including a significant reduction in median error and a significant difference between pre- and post-optimization error distributions (Fig. 6). Improvements are more pronounced in the dHCP dataset and more limited in the Zurich cohort, likely reflecting differences in image quality and segmentation accuracy, consistent with Fig. 9, where reduced corpus callosum visibility is associated with increased variability in both manual and automated landmark placement. These findings suggest that explicitly isolating the corpus callosum could further improve performance.

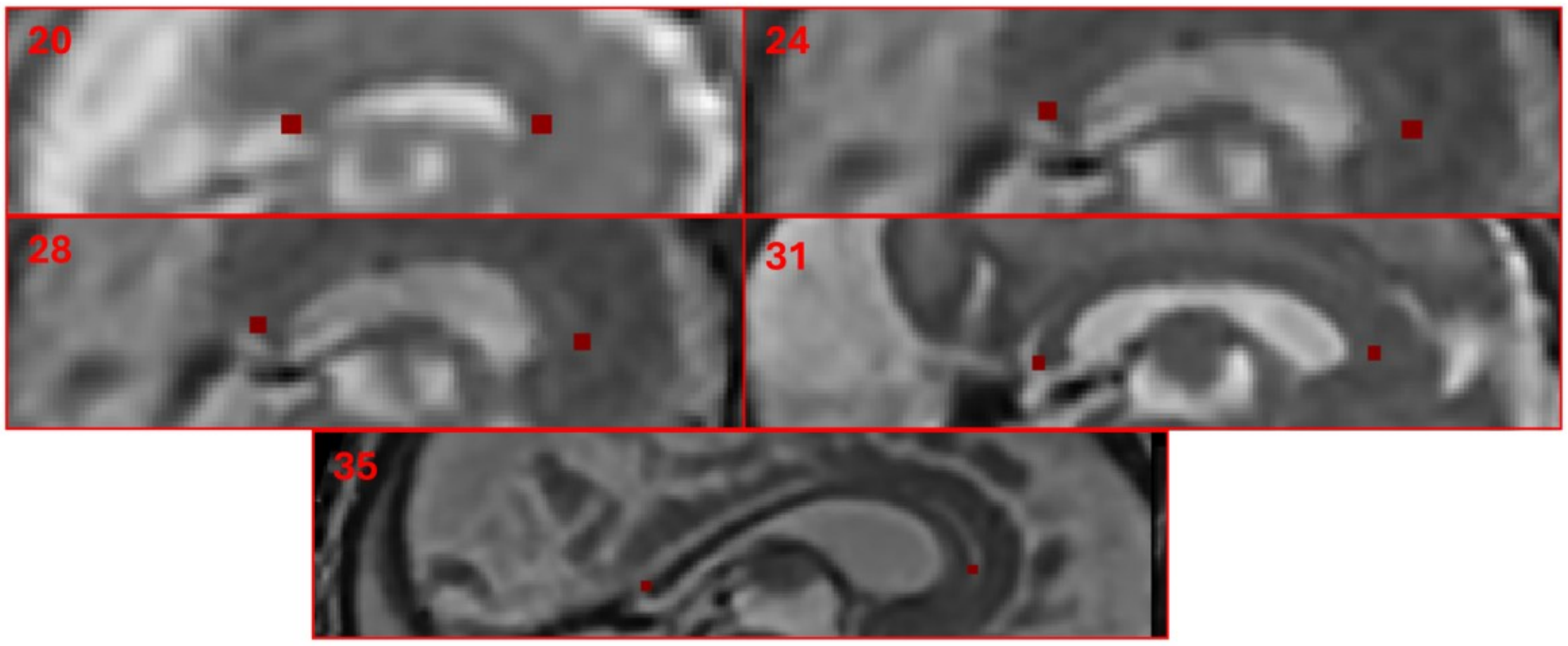

**Fig. 9.** *Corpus callosum landmarks variability. Visualization of manually placed landmarks corresponding to LCC across four different gestational ages from the Zurich dataset (20 and 24 weeks in the first row, 28 and 31 weeks in the second row, and 35 in the third row). Landmarks (in red) had been dilated for visualization purposes. It is possible to appreciate the difficulty in the manual positioning of these landmarks due to the complexities at identifying the corpus callosum margins.*

Automatic landmark detection is a well-established task in CV, with broad applications in medical image analysis [44,45]. In fetal MRI, automated biometry methods estimate measurements either by direct regression from global image features or via predicted anatomical landmarks (Table 1). However, existing studies are limited by the lack of publicly available code and the absence of systematic, quantitative evaluation of landmark localization error, which directly affects clinicians' confidence in automated measurements. For example, [12] report only mean and standard deviation of Euclidean distance errors for measurement endpoints, while [17] assess measurement reliability without explicitly evaluating landmark accuracy. Similarly, [3] rely on expert-based quality scores of landmark positioning. To our knowledge, our study is the first to address these limitations through a systematic, quantitative evaluation of landmark localization accuracy, enabling a more complete characterization of performance beyond measurement accuracy alone. Landmark localization results are reported in Fig. 7 and Table 5. Endpoint pairs show non-identical distributions, reflecting brain asymmetry. Most landmarks achieve a median Euclidean error below 4 mm, indicating good localization accuracy. Errors are lower for smaller structures, such as the vermis, where precise localization is critical, and increase with anatomical size and spatial variability, particularly for bBIP and sBIP, whose landmarks lie on lateral brain and skull boundaries. These regions exhibit greater inter-subject variability, leading to broader error distributions and higher median errors, whereas midline structures are more stable within the reconstructed 3D brain. This tendency is consistent with previous reports [3,12]. Axis-wise analysis (Fig. 7, right) shows that errors for bBIP and sBIP are dominated by the z-axis, indicating that the main challenge lies in identifying the correct imaging plane rather than in-plane localization. This likely reflects variability in acquisition orientation and fetal position, as well as the larger spatial extent of these structures across slices. These observations highlight the advantage of 3D SR reconstruction over conventional 2D slice-based measurements, which are highly sensitive to off-axis plane selection caused by fetal or maternal motion. Operating in 3D mitigates these limitations and enables more reliable biometric estimation [13]. The 3D visualizations (Fig. 8) complement the quantitative analysis by providing intuitive insight into landmark placement and facilitating identification of failure modes beyond numerical metrics.

Our pipeline is straightforward, requiring only good quality segmentation masks, which can be obtained using automated tools when manual annotations are unavailable. Owing to its design, it can readily be scaled to different label maps, enabling additional or alternative measurements. The method is also computationally efficient, requiring approximately 64 minutes for training on a mid-range GPU and only a few seconds per subject for optimization and inference. These characteristics may support integration into clinical workflows, reducing the burden of manual measurements [46].

Despite these promising results, some limitations remain. Expanding the dataset to include a broader range of subjects, scanner manufacturers, acquisition parameters, GAs, and pathological conditions, as well as testing on other external datasets, would further improve robustness and generalizability. Future work should also include additional measurements, such as the fronto-occipital diameter, atrial diameters, and brainstem-related metrics, which are relevant for GA estimation and posterior fossa assessment [47].

In conclusion, this work advances automated fetal brain MRI biometry by introducing a pipeline that enables reliable, reproducible, and interpretable biometric measurements with anatomically consistent landmark localization. Its foundation on 3D SR reconstructions enables true volumetric assessment and supports reliable measurements even for small brain structures.

**Ethical approval**

All data were obtained from publicly available research repositories and were fully de-identified prior to access. Appropriate participant consent and ethical approval were obtained by the original data-collecting studies, in accordance with their respective data use agreements.

**Conflict of interest**

The authors declare that they have no known competing financial interests or personal relationships that could have appeared to influence the work reported in this paper.

**Acknowledgments**

The dHCP data were provided by the developing Human Connectome Project, KCL-Imperial-Oxford Consortium funded by the European Research Council under the European Union Seventh Framework Programme (FP/2007-2013) / ERC Grant Agreement no. [319456]. We are grateful to the families who generously supported this trial.

This work was partially supported by grants from the Italian Ministry of Health to DP ("Ricerca corrente" funds; "Piano Nazionale Complementare Ecosistema Innovativo della Salute - PNC-EJ-2022-23683266 PNC-HLS-DA" funds) and by donation of UniverLecco.

**Data and Code availability**

A code repository on GitHub (https://github.com/Franca-exe/Fetal-brain-biometry) as well as a docker image of the whole pipeline will be openly available to allow reproducibility and foster research in the field (see also Supplementary section S3). Datasets used in the study are publicly available.

**Supplementary**

Table S1

**Table S1.** *Mapping of the 19 tissue labels produced by BOUNTI to the 8-tissue label maps used as input for our experiments. The first column reports the combinations (∪) of BOUNTI labels (denoted as 'label_x', where x corresponds to the label index output by BOUNTI pipeline). The second column indicates the resulting label in the final label map, together with the corresponding anatomical tissue.*

| *BOUNTI labels combination* | *Final label (anatomical tissue)* |
|---|---|
| label_1 ∪ label_2 | *1 (CSF)* |
| label_3 ∪ label_4 | 2 (GM) |
| label_5 ∪ label_6 | 3 (WM) |
| label_7 ∪ label_8 ∪ label_9 ∪ label_18 ∪ label_19 | 4 (ventricles) |
| label_11 ∪ label_12 ∪ label_13 | 5 (cerebellum) |
| label_14 ∪ label_15 ∪ label_16 ∪ label_17 | 6 (deep GM) |
| label_10 | 7 (brainstem) |

Figure S1

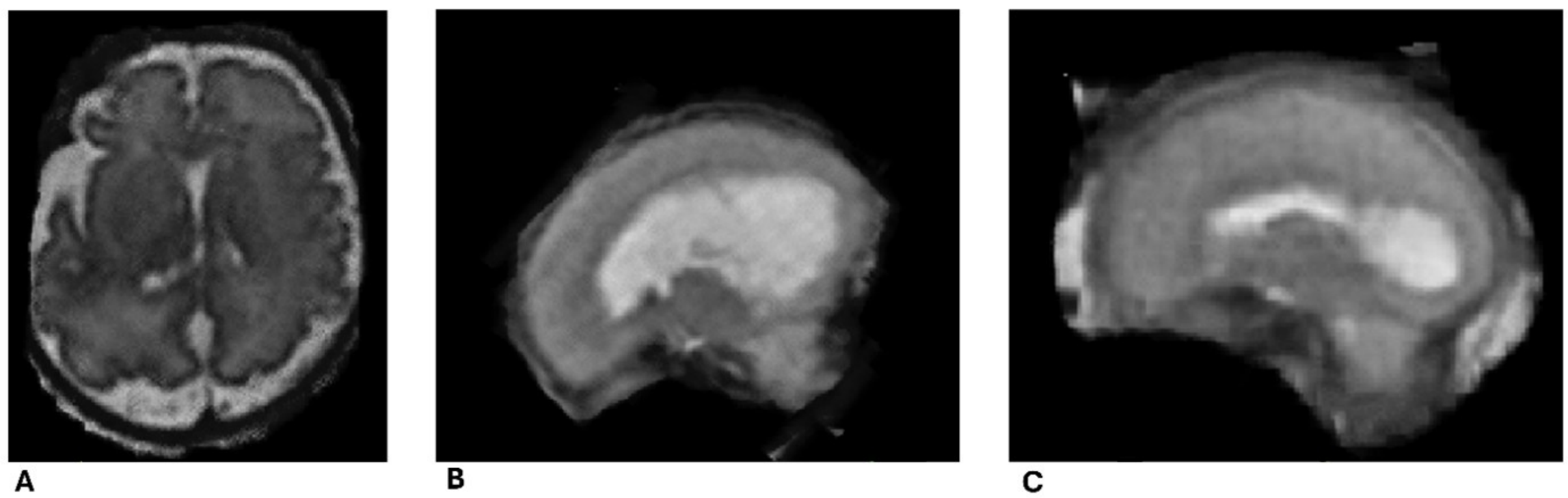

**Fig. S1**. *Examples of three subjects from the Zurich dataset in which acquisition-related noise limits the landmarking process. Acquisitions are represented according to the best view showing the limitation to the landmarking procedure. Subject A (GA = 32.8 weeks) lacks landmarks associated with the sBIP measurement; Subject B (GA = 24.2 weeks) lacks cerebellar landmarks, specifically those required for TCD and HV measurements; and Subject C (GA = 23.3 weeks) lacks landmarks associated with the LCC, sBIP, and bBIP measurements. Due to image noise, annotations in the Zurich cohort are incomplete: 10 subjects lack annotations entirely, and 6 have partial annotations (specifically, 3 subjects have 1 pair of landmarks missing, 1 subject has 2 pairs of landmarks missing, and one has one pair of landmarks available).*

Figure S2

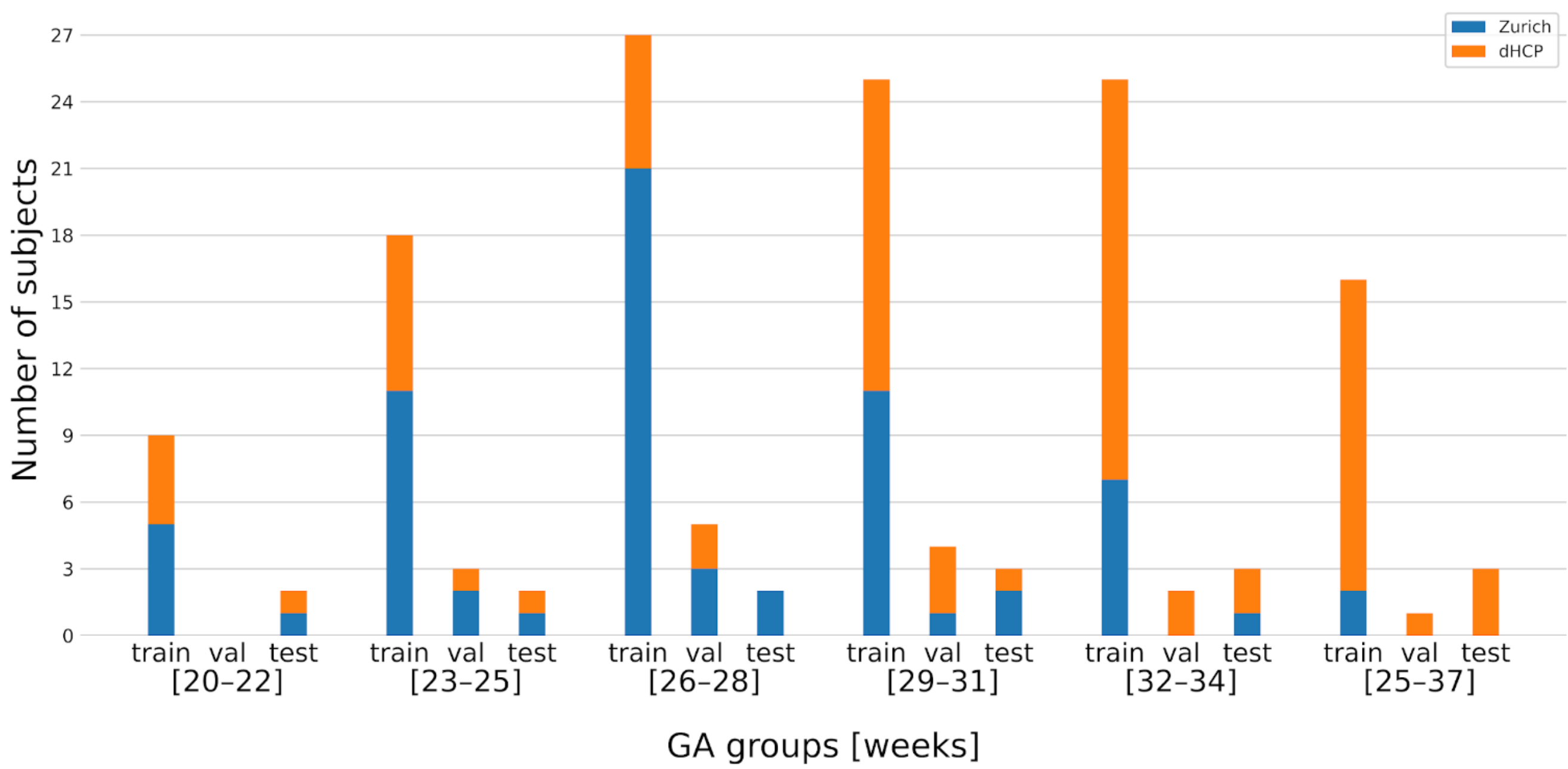

**Fig. S2**. *Data split across the two datasets and grouped by GA ranges. We group the data into six GA ranges ([20-22], [23-25], [26-28], [29-31], [32-34], [35-37] weeks) and consider the ratio between the two datasets. We aim to keep this proportion as consistent as possible across the three subsets to reflect the overall distribution of the dataset. For each GA group, each bar shows the cumulative number of samples in the training ('train'), validation ('val'), and testing ('test') sets splitted into subjects from the Zurich dataset (blue) and from the dHCP dataset (orange).*

**S1. Ground-truth vs BOUNTI labels**

Our method is trained using anatomical segmentation label maps. In our experiments, we try two different types of input label maps given by two different sources. We consider label maps that are provided with the dataset, which we consider and refer to as ground-truth, as well as the segmentation masks obtained with a third-party software that is considered the state-of-the-art for fetal brain tissue segmentation, namely BOUNTI. Since we want to test the robustness of our method with respect to the input, namely the segmentation masks provided to the network, we compare the effect of the segmentation on the output results. Therefore, we conduct experiments considering both GT and BOUNTI masks for the network training and optimization process. We compare the distribution of absolute error and percentage error computed on the predicted measurements across the test set, as represented by Fig. S3. For both types of error distributions, paired t-test revealed that there is no significant difference between the distributions using the two types of labels, except for LCC (p-value=0.016 for absolute error and p-value=0.015 for the percentage error).

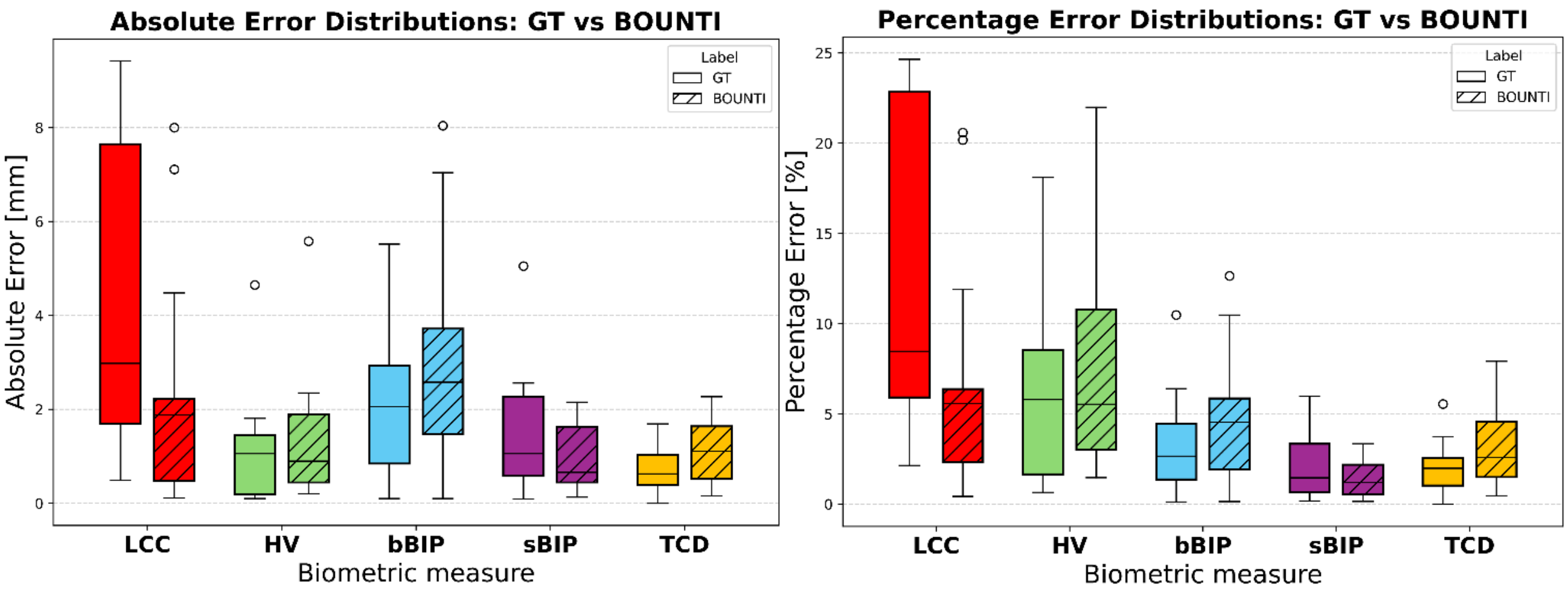

**Fig. S3**. *Error distributions across test set subjects considering ground truth or BOUNTI label maps. We report the absolute and percentage error distributions for each biometric measurement obtained considering two possible cases: one setting in which we provide to the network anatomical segmentations considered as the ground-truth ones (plain boxes), namely provided with the dataset, and the second one in which we exploit segmentations coming from the third-party software BOUNTI (hatched boxes).*
*Acronym: GT: ground-truth.*

**S2. k-parameter setting**

Our pipeline (see subsection 2.1.1) requires the transformation of input to the CNN to the atlas space (Uus et al., 2023) leading to a volume being aligned to this space at 0.5 mm isotropic resolution. Since landmarks images are sparse volumes in which each landmark occupies a single voxel, during the registration process, due to image interpolation and resampling, the landmark information can be lost. To avoid this information loss, we apply a dilation of landmarks via a morphological operation using a structuring element, whose dimension is defined by the k parameter. We start from a heuristically predefined value of k set to 5, which assures landmarks are preserved after registration. In our experiments, we lowered this parameter until reaching the minimum value for which landmarks are preserved, which is found out to be equal to 3.

**S3. Technical details**

To prepare data and train our experiments, we have defined an Anaconda environment with Python version 3.10.12. We use the following main Python packages: Pytorch v.2.2.2 and torchvision v.0.17.2 with CUDA v.11.8, torchaudio v.2.2.2, SimpleITK v.2.3.1, pandas v.2.1.4, antspyx v.0.4.2 , nibabel v.5.2.1, nilearn v.0.10.4, numpy v.1.24.4, lightning v.2.3.0, lightning-utilities==0.11.6, pytorch-lightning v.2.3.0, fslpy v.3.21.0, tensorboardX and tensorboard.
Other softwares that we used during data preparation are: ITK-SNAP (v.3.6.0), ANTs (v.2.5.3), and FSL (v.5.0).
Training experiments are performed using a laptop with NVIDIA GeForce RTX 2050 GPU.

Metrics computations and statistical analysis are performed using both Microsoft Excel and its add-in Statistical analysis and in Python utilizing appropriate libraries as SciPy and Pingouin.